%% file: main.tex
\documentclass[10pt,twocolumn,letterpaper]{article}

\usepackage{wacv}
\usepackage{times}
\usepackage{epsfig}
\usepackage{graphicx}
\usepackage{amsmath}
\usepackage{amssymb}

\wacvfinalcopy % *** Uncomment this line for the final submission

\def\wacvPaperID{616} % *** Enter the wacv Paper ID here

\ifwacvfinal\fi
\begin{document}

%%%%%%%%% TITLE
\title{PointGrow: Autoregressively Learned Point Cloud Generation with Self-Attention}

% % Authors at the same institution
% \author{Yongbin Sun \hspace{2cm} Yue Wang \\
% Massachusetts Institute of Technology\\
% {\tt\small yb\_sun@mit.edu,  yuewang@csail.mit.edu}
% }
% Authors at different institutions
% \author{
% Yongbin Sun \\
% Massachusetts Institute of Technology\\
% {\tt\small  yb\_sun@mit.edu}
% \and
% Yue Wang \\
% Massachusetts Institute of Technology\\
% {\tt\small yuewang@csail.mit.edu}
% \and
% Ziwei liu \thanks{corresponding author.}\\
% The Chinese University of Hong Kong\\
% {\tt\small zwliu.hust@gmail.com}
% \and
% Joshua E. Siegel \\
% Michigan State University\\
% {\tt\small jsiegel@msu.edu}
% \and
% Sanjay E. Sarma \\
% Massachusetts Institute of Technology\\
% {\tt\small sesarma@mit.edu}
% }

\author{Yongbin Sun$^1$,~~~Yue Wang$^1$,~~~Ziwei Liu$^2$\thanks{corresponding author.},~~~Joshua E Siegel$^3$~~\&~~Sanjay E Sarma$^1$\\ \\
1. Massachusetts Institute of Technology~~2. The Chinese University of Hong Kong\\
3. Michigan State University
}

\maketitle
\ifwacvfinal\thispagestyle{empty}\fi

\input{./sections/abs.tex}
\input{./sections/intro.tex}
\input{./sections/related_work.tex}
\input{./sections/approach.tex}

\input{./sections/exp.tex}

\input{./sections/conclusion.tex}

{\small
\bibliographystyle{./ieee}
\bibliography{./egbib}
}

\end{document}

%% file: sections/abs.tex
\begin{abstract}
% A point cloud is a 3D representation for efficiently modeling an object's surface geometry. 
% Clouds' surface-centric properties pose challenges to designing recognition and synthesis tools. 
Generating 3D point clouds is challenging yet highly desired. 
This work presents a novel autoregressive model, \emph{PointGrow}, which can generate diverse and realistic point cloud samples from scratch or conditioned on semantic contexts. 
This model operates recurrently, with each point sampled according to a conditional distribution given its previously-generated points, allowing inter-point correlations to be well-exploited and 3D shape generative processes to be better interpreted. 
Since point cloud object shapes are typically encoded by long-range dependencies, we augment our model with dedicated self-attention modules to capture such relations.
Extensive evaluations show that \emph{PointGrow} achieves satisfying performance on both unconditional and conditional point cloud generation tasks, with respect to realism and diversity.
Several important applications, such as unsupervised feature learning and shape arithmetic operations, are also demonstrated.
% Further, \emph{conditional PointGrow} learns a smooth manifold of given image conditions inside of which 3D shape interpolation and arithmetic calculation can be performed. 
\end{abstract}

%% file: sections/intro.tex
\section{Introduction}

% Shape generation is important.
% Understanding 3D shape generative process has always been at the core of computer vision, and is still an open challenge today. 
Recently 3D generative model has attracted enormous research interests because it directly promotes the development of emerging applications, such as virtual/augmented reality \cite{stets2017visualization, sun2018x} and self-driving cars.
For example, it is capable of completing the LIDAR scans that might suffer from occlusion issues \cite{yuan2018pcn}. 
% Additionally, 3D generative modeling can also benefit resolving discriminative tasks (e.g. classification and segmentation) \cite{qi2017pointnet, qi2017pointnet++, wang2018dynamic} by generating 3D shapes for training discriminative models, which will benefit more applications, including autonomous  vehicles \cite{yue2018lidar}, indoor navigation \cite{diaz2016indoor} and robotics \cite{varley2017shape}, etc.
% These benefits are becoming even more significant today, considering that the popular deep learning-based approach enjoys a data-driven training process and usually requires training data at a large scale, but manually collecting real 3D shapes using LIDARs or depth cameras is time consuming and designing synthetic 3D shapes requires additional background and expertise of 3D design. 
Therefore, intelligent systems that are able to automatically generate realistic and diverse 3D shapes are highly desired.
% , which forms the focus of this paper.

% Other types of shape generation: discussion.
3D shapes are usually represented as triangle meshes or point clouds due to their light-weight nature and simple form.
Such shape representations are flexible for rendering, but impose a problem when applying computer vision techniques for processing, because most standard operations are designed based on regular grid-based formats (e.g. image), while meshes and point clouds are fundamentally irregular: vertex or point positions are continuously distributed in the space, and any permutation of their face or point ordering does not change the spatial distribution.
Therefore, one major line of 3D research discretizes a continuous shape representation onto a 3D grid \cite{brock2016generative, maturana2015voxnet, wu2016learning, wu20153d}.
But such volume-based representations consume significant memory and introduce quantization artifacts, making it difficult to generate high-res 3D shapes and retain fine-grained surface details.
Therefore, it is more desirable to design a framework specific to raw shape representations, rather than relying upon intermediate representations. 
In this paper, we choose to represent shapes using point clouds, because they comprise the output of most existing 3D sensing technologies, showing more application values but with less complexity.

% Discuss existing point cloud based shape generative process, leading to what we want to do -> why we model shape distribution -> autoregressive model.
The vast majority of existing learning-based works for 3D point clouds generation rely on two types of distance metric between point sets, Chamfer Distance (CD) and Earth Mover's Distance (EMD) \cite{fan2017point}, to handle the irregularity problem of point clouds.
Serving as the loss function, these distance metric penalizes the dissimilarity between the generated and ground truth point sets, and help deep generative models learn to produce visually-plausible 3D shapes for various types of inputs \cite{gadelha2018multiresolution, lin2017learning, achlioptas2018learning, gadelha2017shape, jiang2018gal, yu2018pu, groueix2018atlasnet}.
However, the generative process is hard to interpret due to the intrinsic limitation of \textit{inter-set} distance metric. 

% Our approach, need to compare with distance-based methods.
In this work, we seek to explore a different approach to better \emph{understand} and \emph{interpret} the point cloud generative process.
We begin by observing that the points constituting a 3D shape have correlations.
For example, most man-made shapes are symmetric, such as the four legs of a table;
also, points are distributed in a correlated way to form different parts of a shape (e.g. to produce an airplane, some points form its wings, while others have to form its body structure).
To explicitly learn and utilize such \emph{inter-point} correlations for shape generation, we adopt a probabilistic approach to jointly model the spatial distribution of all the points of a shape in an $n$ dimensional space, where $n$ is the number of points in a point cloud.
Underpinned by a joint point distribution reflecting the underlying inter-point correlation in the data, a high joint probability value should correspond to the point set distribution of a plausible 3D shape, while a low value should indicate an implausible one.
%In this way, the joint point distribution associated to a plausible 3D shape should have a high probability value, while a low probability value should indicate the point set of an implausible one. 
%With a joint distribution reflecting the underlying inter-point correlation in the data available, the point cloud generation process can be then cast as a sampling process in an $n$ dimensional point space.
In this way, the point cloud generation process can be cast as a sampling process in an $n$ dimensional point space.

\begin{figure}[t!]
% \vspace{-40pt}
\begin{center}
\includegraphics[width=1.0\linewidth]{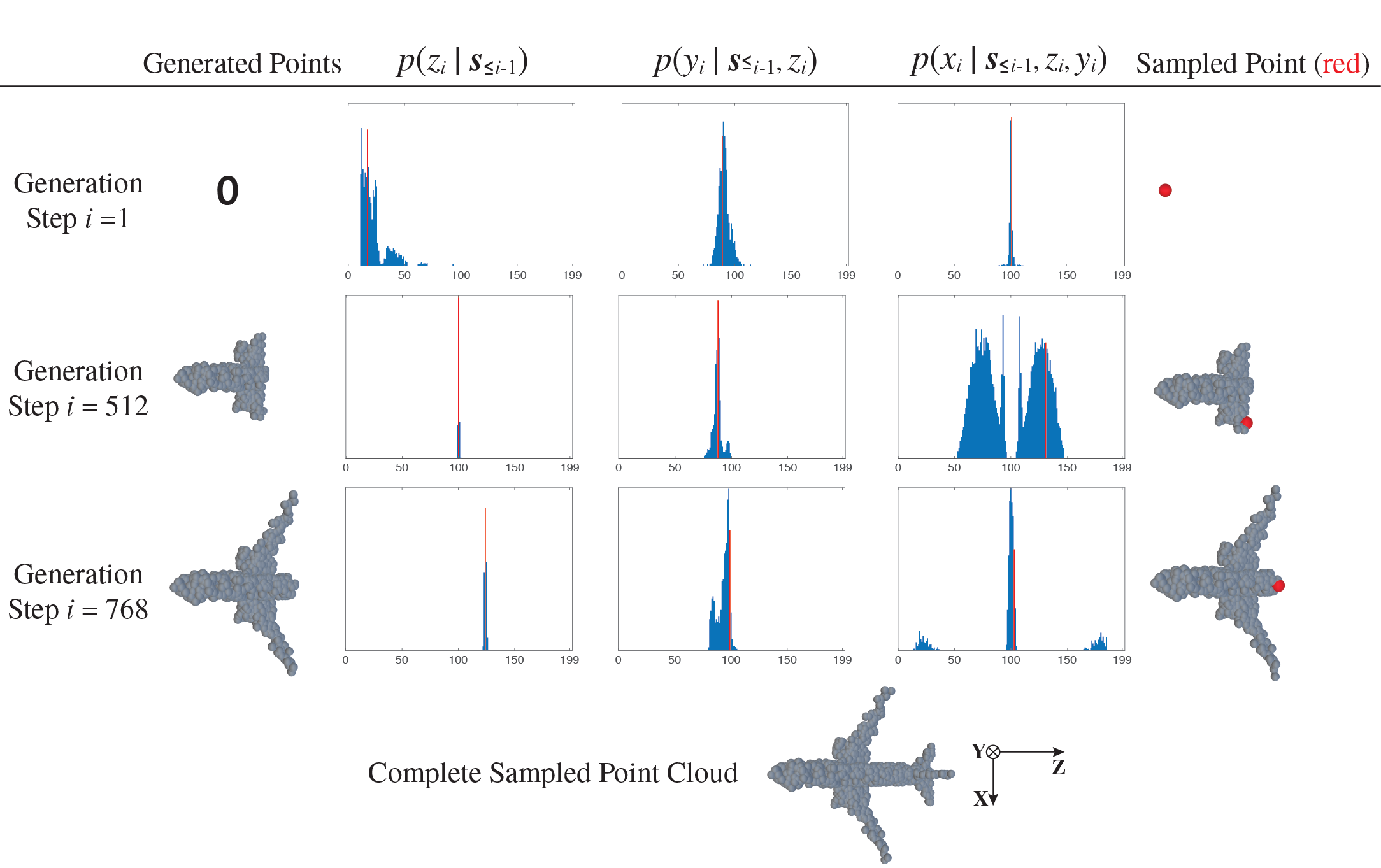}
\end{center}
\vspace{-10pt}
\caption{The point cloud generation process in \emph{PointGrow} (best viewed in color). Given $i-1$ generated points, our model estimates a conditional distribution of coordinate $z_i$, indicated as $p(z_i|\textbf{s}_{\leq i-1})$, and then samples a value (indicated as a red bar) according to this distribution. The process is repeated to sample $y_i$ and $x_i$ with previously sampled coordinates as additional conditions. The $i^{th}$ point (red point in the last column) is obtained as $\{x_i, y_i, z_i\}$. Note that $x_{512}$ shows a symmetric wing shaped conditional distribution.}
\label{fig:prog}
\vspace{-10pt}
\end{figure}

Furthermore, since a joint probability can be decomposed by chain rule as the product of a series of conditional probabilities, the joint point set probability can be  expressed by a series of conditional point probabilities, where each point is conditioned on its previously generated ones.
This property naturally enables us to visualize the shape generative process and interpret inter-point correlations in a \emph{point-by-point} manner during the point sampling process, as shown in Figure \ref{fig:prog}.

To this end, we propose an autoregressive framework dubbed \emph{PointGrow} to generate every point recurrently.
Specifically, \emph{PointGrow} estimates a conditional distribution of the point under consideration given all its preceding points. 
However, the irregularity of point clouds imposes difficulties when aggregating meaningful information from a given point set, especially when such information is contained by distant points.
Therefore, we further propose two point cloud-based self-attention modules to dynamically aggregate long-range dependencies from available points. 
Our experiments show that those two modules can improve information flow between points, and successfully capture meaningful semantic information.
% In addition, the \emph{conditional PointGrow} learns a smooth manifold of given images, where interpolation and arithmetic calculation can be performed on image embeddings.
% Another benefit provided by autoregressive models is that a simple cross-entropy loss handles the shape learning process efficiently without requiring the use of a computationally-intensive set-to-set distance loss function.

The contributions of our work are summarized as below:
\begin{itemize}
    % \item{We successfully model the joint spatial distribution of 3D shape points in an autoregressive manner, such that the 3D shape generative process can be easily understood and interpreted.}
    \item{We propose a novel autogressive model, \emph{PointGrow}, for point cloud generation, which models the joint 3D spatial distribution in a \emph{point-by-point} manner. PointGrow has two appealing properties: 1) it is capable of generating diverse and realistic 3D clouds, 2) it constitutes an interpretable 3D shape generative process.}
    
    \item{Two self-attention modules are carefully designed to capture long-range dependencies and semantic correlations between points, facilitating the generation of plausible part configurations within 3D objects.}
    
    \item{Besides point cloud generation, our framework also enables several important applications, such as diverse shape completions, unsupervised feature learning and shape arithmetic operations.}
\end{itemize}

%% file: sections/related_work.tex
\section{Related Work}
While 3D data processing and generation has a long history, here, we only discuss the directly related work of using deep networks to analyze 3D shapes, autoregressive networks, and self-attention. 

\subsection{Shape Analysis}
%%%%%%%%%%%%%%%%%%%%%%%%%%
\noindent
\textbf{Volumetric Methods.}
3D shape recognition and generation has been studied using 3D voxel grids \cite{brock2016generative, choy20163d, maturana2015voxnet, wu2016learning, wu20153d, sun2018im2avatar}. 
Voxelization often produces a sparsely-occupied 3D grid, which limits resolution and introduces quantization artifacts.
Recent frameworks have been proposed to reduce spatial complexity of volumetric shape representations, \cite{graham2017submanifold, hane2017hierarchical, riegler2017octnet, su2018splatnet, tatarchenko2017octree, wang2017cnn}, though these generally suffer from high computation costs.  

%%%%%%%%%%%%%%%%%%%%%%%%%%%
% Figure
%%%%%%%%%%%%%%%%%%%%%%%%%%%
\begin{figure*}[t!]
\vspace{-20pt}
\begin{center}
\includegraphics[width=0.95\linewidth]{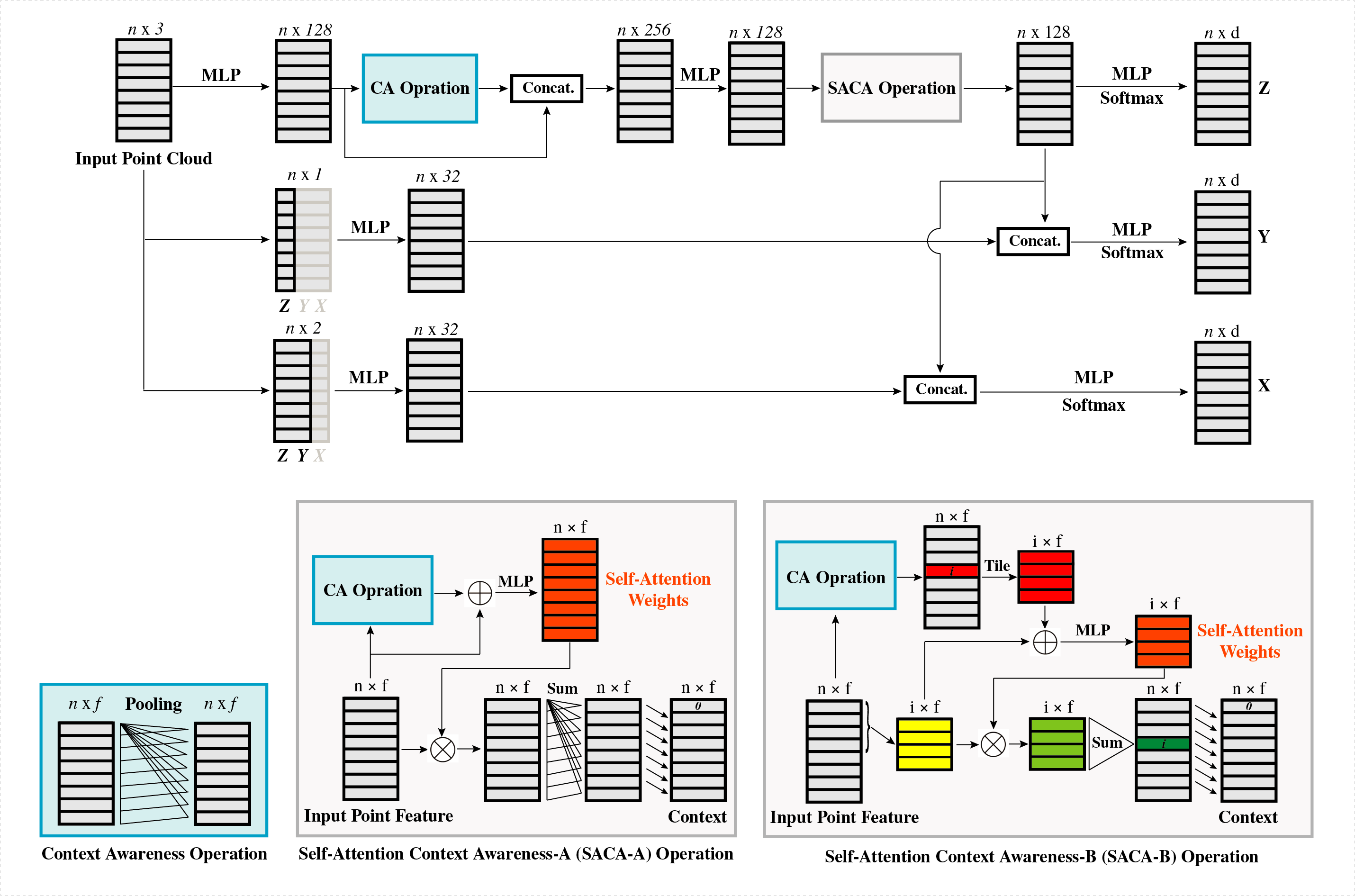}
\end{center}
\vspace{-15pt}
\caption{The proposed model architecture and context awareness operations to estimate conditional distributions of point coordinates. $\otimes$: element-wise production, $\oplus$: concatenation.}
\label{fig:ops}
\vspace{-15pt}
\end{figure*}
%%%%%%%%%%%%%%%%%%%%%%%%%%%

% \noindent
% \textbf{View-Based Methods.}
% View-Based methods represent a 3D shape using a collection of rendered 2D views, allowing standard CNNs to be applied \cite{kalogerakis20173d, lun20173d, qi2016volumetric, su2015multi}.
% 2D projected views describe exterior surfaces well, lending such methods to discriminative tasks. 
% \cite{soltani2017synthesizing, lin2017learning} recently demonstrated 3D shape synthesis using image-based networks, but additional processing is required to address view inconsistencies and self occlusion.

\noindent
\textbf{Mesh-Based Methods.}
3D meshes are a lightweight approach for geometric modelling via a set of vertices and triangular or quad primitives.
Recent work has extended standard convolutions to mesh surfaces for aggregating and propagating local features \cite{boscaini2016learning, bruna2013spectral, masci2015geodesic, yi2017syncspeccnn}.
Relevant work reconstructing 3D shapes as 3D meshes is found in \cite{wang2018pixel2mesh, achlioptas2018learning, sengupta2017sfsnet, kanazawa2018learning, kong2017using, pontes2017compact, pontes2017image2mesh, jack2018learning, tan2017variational}. 

\noindent
\textbf{Point Cloud-Based Methods.}
PointNet \cite{qi2017pointnet} is the pioneering work in applying deep neural nets to point sets, using a symmetric function to aggregate feature vectors for all points in a permutation-invariant manner. 
PointNet's successors explore ways to accumulate local information in the spatial \cite{qi2017pointnet++, shen2017neighbors, klokov2017escape, qi2017frustum} and embedded feature domains \cite{wang2018dynamic, li2018pointcnn, te2018rgcnn} to achieve high performance. 
To address point cloud generative tasks, \cite{fan2017point} introduced two symmetric distance metrics, CD and EMD, to measure the distance between two point sets. 
These metrics are order-invariant, which makes them suitable as loss function operated directly on point clouds.
By taking advantage of these metrics, models have been proposed to address point cloud synthesis problems under different settings \cite{gadelha2018multiresolution, lin2017learning, achlioptas2018learning, gadelha2017shape, jiang2018gal, yu2018pu, yang2018foldingnet, groueix2018atlasnet, yuan2018pcn}.
However, existing generative approaches focus on measuring inter-set dissimilarity, and the inter-point relationship within a point set is not well understood.

%%%%%%%%%%%%%%%%%%%%%%%%%%%
%%%%%%%%%%%%%%%%%%%%%%%%%%%
\subsection{Autoregressive Networks}
Autoregressive networks model current values as a function of their own previous values, and have been adopted to model the joint distribution of image pixels \cite{oord2016pixel, van2016conditional, salimans2017pixelcnn++} and audio samples \cite{van2016wavenet}.
The joint distribution is cast as a product of conditional distributions, and each condition distribution is modeled using a deep neural network that takes as input previously generated values and outputs a distribution for the value currently under consideration.
But it is not trivial to adapt autoregressive frame to point cloud due to the irregularity problem of point clouds.
% The autoregressive network approach addresses our goals of exploring and understanding inter-point relations in point cloud generative processes well, though careful design is required to handle intrinsic properties of point clouds while allowing efficient information propogation. 

%%%%%%%%%%%%%%%%%%%%%%%%%%%
%%%%%%%%%%%%%%%%%%%%%%%%%%%
\subsection{Self-Attention}
Attention is a flexible mechanism to capture information in a self-adaptive manner such that accumulated important information is weighted highly.
It improves performance in tasks including image recognition \cite{larochelle2010learning, denil2012learning} and natural language processing \cite{cheng2016long, vaswani2017attention, bahdanau2014neural}.
Recently, self-attention has been adopted into generative tasks, such as image generation \cite{zhang2018self}.
% \cite{zhang2018self} shows that fine local visual details can be generated after considering distant inter-pixel relations. %% TODO: Not sure what you're trying to say here
In our experiments, we demonstrate that self-attention modules can be extended to process unordered point sets and capture inter-point correlations.

%% file: sections/approach.tex
\section{PointGrow}
This section introduces the formulation and implementation of \emph{PointGrow} (and its conditional version), a point-by-point generative model for 3D point clouds.

% %%%%%%%%%%%%%%%%%%%%%%%%%%%
% % Figure
% %%%%%%%%%%%%%%%%%%%%%%%%%%%
% \begin{figure*}[t!]
% \vspace{-20pt}
% \begin{center}
% \includegraphics[width=0.9\linewidth]{./sections/figures/point_gen.png}
% \end{center}
% \vspace{-10pt}
% \caption{Self-attention fields in \emph{PointGrow} for query locations (red points), visualized as Euclidean distances between the context feature of a query point and the features of its accessible points. Note that points highly correlated to the query point are not necessarily spatially-close. \textbf{Left:} SACA-A; \textbf{Right:} SACA-B.}
% \label{fig:attention_map}
% \vspace{-15pt}
% \end{figure*}
% %%%%%%%%%%%%%%%%%%%%%%%%%%%
% %%%%%%%%%%%%%%%%%%%%%%%%%%%

%%%%%%%%%%%%%%%%%%%%%%%%%%%
\noindent
\textbf{Unconditional PointGrow.}
A point cloud, $\textbf{S}$, that consists of $n$ points is defined as $\textbf{S} = \{\textbf{s}_{1}, \textbf{s}_{2}, ..., \textbf{s}_{n}\}$, with its $i^{th}$ point $\textbf{s}_{i} = \{x_i, y_i, z_i\}$ in 3D space. 
Our goal is to assign a probability $p(\textbf{S})$ to each point cloud.  
We do so by factorizing the joint probability of $\textbf{S}$ as a product of conditional probabilities over all its points:
\begin{equation}
    p(\textbf{S}) = \prod_{i=1}^{n} p(\textbf{s}_i|\textbf{s}_1,...,\textbf{s}_{i-1}) = \prod_{i=1}^{n} p(\textbf{s}_i|\textbf{s}_{\leq i-1})
\label{eq:1}
\end{equation}
The value $p(\textbf{s}_i|\textbf{s}_{\leq i-1})$ is the conditional probability of the $i^{th}$
point $\textbf{s}_{i}$ given all its previously generated points, and computed as a joint probability over its coordinates:
\begin{equation}
    p(\textbf{s}_i|\textbf{s}_{\leq i-1}) = p(z_i|\textbf{s}_{\leq i-1}) \cdot p(y_i|\textbf{s}_{\leq i-1}, z_i) \cdot p(x_i|\textbf{s}_{\leq i-1}, z_i, y_i),
\label{eq:2}
\end{equation}
where each coordinate is conditioned on available coordinates.
To facilitate the generation process, we sort training points according to their $z$ coordinates to encourage a shape to be generated mainly along its primary axis during testing (like 3D printing), hoping that semantic information can be better captured to produce consistent shapes (e.g. a rear car body to be generated has to match an existing front).
But since our model is sampling-based, the generated coordinates are not strictly larger than their previous ones and processing modules (introduced later) should be invariant to point permutation. 
% To facilitate point cloud generation, we sort points in the order of $z$, $y$ and $x$, which forces a shape to be generated in a ``plane-sweep" manner along its primary ($z$) axis.
Here, we model the conditional probability distribution of each coordinate using a deep neural network. 
Prior art \cite{oord2016pixel} shows that a softmax discrete distribution is more flexible than a continuous one to model any arbitrary distribution.
So we discrete point coordinates by scaling them into the range [0, 1] and quantizing them to $d$ uniformly distributed values.
Note that different from many existing voxel-based methods generating tensors of size $d^3$ and operating on 3D volumes, our model outputs tensors of $3 \times n \times d$ (usually much smaller than $d^3$ considering the resolution of shape volumes) and operates directly on sparse point representation.
We set $n=1024$ and $d=200$ as a trade-off between generative performance and quantization artifacts. 
Larger $n$ and $d$ can be used to achieve better visual results but at the cost of slower performance.
% Other advantages of adopting discrete coordinates include (1) simplified implementation, (2) improved flexibility to approximate any arbitrary distribution, and (3) it prevents generating distribution mass outside of the range, which might occur in continuous cases. 

%%%%%%%%%%%%%%%%%%%%%%%%%%%%%%%%%%%%%%%%%%
\noindent
\textbf{Context Awareness Operation.}
Context awareness improves model inference.
For example, in \cite{qi2017pointnet} and \cite{ wang2018dynamic}, a global feature is obtained by applying max pooling along each feature dimension, and then used to provide context information for solving semantic segmentation tasks.
Similarly, we obtain context-aware features for all sets of generated available points in the point cloud generation process, as illustrated in Figure \ref{fig:ops} (bottom left).
Each row of resultant context-aware features aggregates the context information of all the previously generated points dynamically by fetching and averaging.
This Context Awareness (CA) operation is implemented as a plug-in module in our model, and mean pooling is used in our experiments.

%%%%%%%%%%%%%%%%%%%%%%%%%%%%%%%%%%%%%%%%%%
\noindent
\textbf{Self-Attention Context Awareness Operation.}
The CA operation accumulates point features in a fixed manner via pooling.
Improving this, we propose two learning-based operations to determine the weights for aggregating point features self-adaptively. 
We define these as Self-Attention Context Awareness (SACA) operations, and the weights as self-attention weights.

Figure \ref{fig:ops} shows the first SACA operation, SACA-A.
To allow each input point feature to understand its importance in context and later determine its weight, we associate it with its context-aware feature obtained after a CA module. 
The combined feature vector is then passed into a Multi-Layer Perception (MLP) to learn self-attention weights.
Given an $n \times f$ point feature matrix, $\textbf{F}$, with its $i^{th}$ row, $\textbf{f}_{i}$, representing the feature vector of the $i^{th}$ point, we compute the $i^{th}$ self-attention weight vector, $\textbf{w}_{i}$, as below:
\begin{equation}
    \textbf{w}_i = MLP(\underset{1 \leq j \leq i}{Mean}\{\textbf{f}_{j}\} \oplus \textbf{f}_{i}),
\vspace{-5pt}
\end{equation}
where $Mean\{\cdot\}$ is mean pooling, $\oplus$ is concatenation, and $MLP(\cdot)$ is a sequence of fully connected layers.
The self-attention weights encode information about context changes due to each newly generated point, and are unique to that point.
Next, we conduct element-wise multiplication between input point features and self-attention weights to obtain weighted features, which are then accumulated sequentially to generate corresponding context features.
The process to calculate the $i^{th}$ context feature, $\textbf{c}_i$, is summarized as:
\begin{equation}
\vspace{-1pt}
    \textbf{c}_i = \sum_{m = 1}^{i} \textbf{f}_m \otimes \textbf{w}_m  = \sum_{m = 1}^{i} \textbf{f}_m \otimes MLP(\underset{1 \leq j \leq \textbf{m}}{Mean}\{\textbf{f}_{j}\} \oplus \textbf{f}_{m}),
\label{eq:saca-a}
\end{equation}
where $\otimes$ is element-wise multiplication.
Finally, we shift context features downward by one row, because for the $i^{th}$ point, $\textbf{s}_i$, only its previous points, $\textbf{s}_{\leq i-1}$, are available.
A zero vector of the same size is attached to the beginning as the initial context feature, indicating no a-priori context knowledge is available.

Figure \ref{fig:ops} also shows the other SACA operation, SACA-B.
SACA-B differs from SACA-A in the way to compute and apply self-attention weights.
In SACA-B, the $i^{th}$ context-aware feature after CA operation is shared by all the first $i$ point features to obtain self-attention weights, which are used to compute $\textbf{c}_i$.
This process is described as:
\begin{equation}
    \textbf{c}_i = \sum_{m = 1}^{i} \textbf{f}_m \otimes \textbf{w}_m = \sum_{m = 1}^{i} \textbf{f}_m \otimes MLP(\underset{1 \leq j \leq \textbf{i}}{Mean}\{\textbf{f}_{j}\} \oplus \textbf{f}_{m})
    \label{eq:saca-b}
\end{equation}{}
Compared to SACA-A, SACA-B self-attention weights encode the importance of each point feature under a common context.
Their differences are highlighted in Eq. (\ref{eq:saca-a}) and (\ref{eq:saca-b}).

\begin{figure*}[t!]
\vspace{-20pt}
\begin{center}
\includegraphics[width=1.0\linewidth]{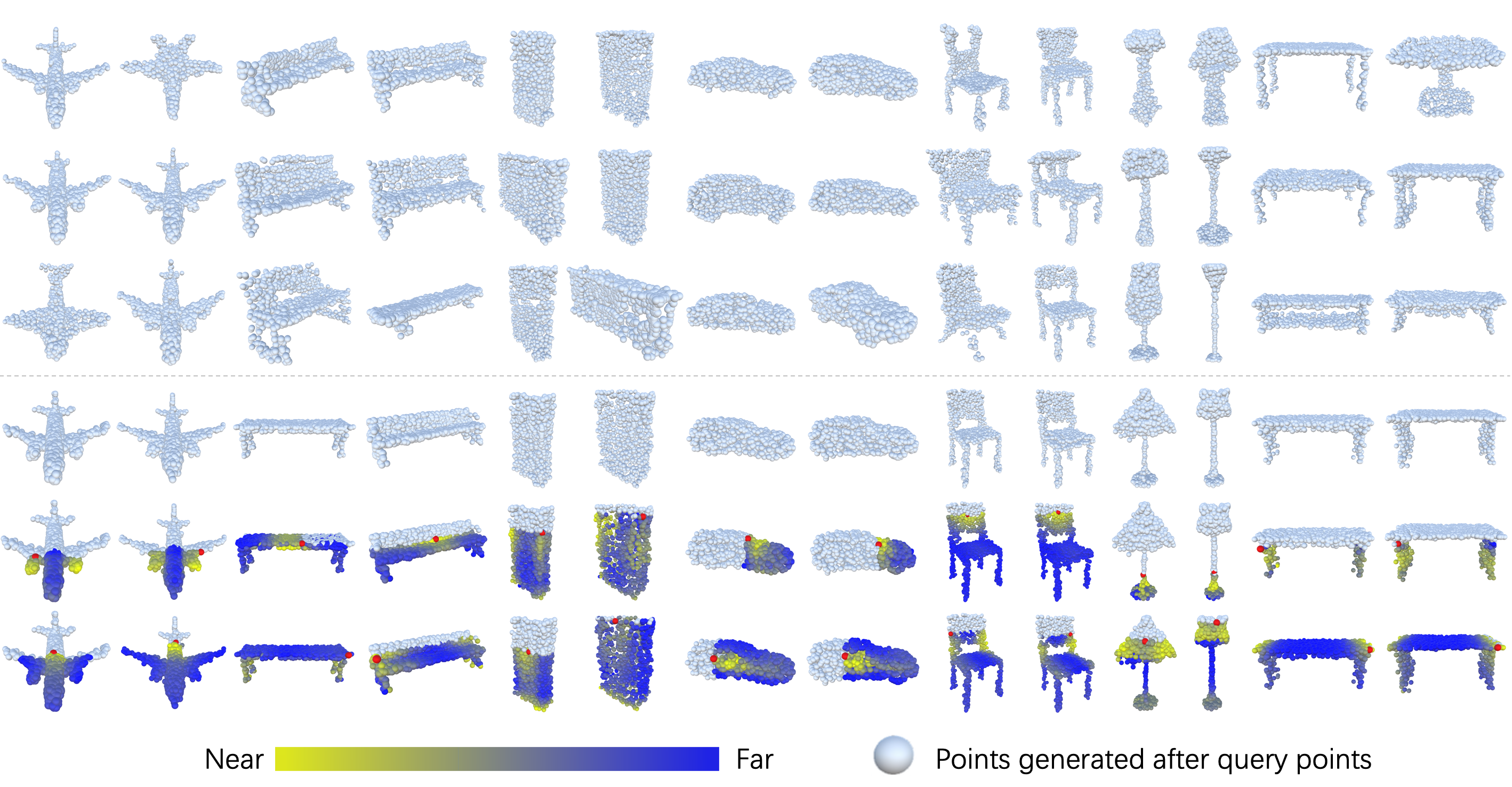}
\end{center}
\vspace{-10pt}
\caption{Visualize generated shapes by \emph{PointGrow} (for each two-column shape set, Left: SACA-A; Right: SACA-B). The bottom part additionally shows self-attention fields, measured by Euclidean distance in feature space, for query locations (red points). Note that the attention modules capture not only spatial but also semantic correlations. For example, when generating wing structure points, the model attends more on other available wing structure points, and some are spatially far but semantically close.}
\label{fig:attention_map}
\vspace{-15pt}
\end{figure*}

% Inside SACA operations, the learning happens only within MLP.
In Figure \ref{fig:attention_map}, we visualize the attention fields during generative processes by visualizing Euclidean distances between the context feature of a query point and the point features before the SACA operation of its previously generated points.

%%%%%%%%%%%%%%%%%%%%%%%%%%%%%%%%%%%%%%%%%%
% \begin{figure}[t!]
% \vspace{-30pt}
% \begin{center}
% \includegraphics[width=1.0\linewidth]{./sections/figures/pipeline.png}
% \end{center}
% \caption{The proposed model architecture to estimate conditional distributions of point coordinates.}
% \label{fig:pipeline}
% \end{figure}

%%%%%%%%%%%%%%%%%%%%%%%%%%%
% Figure
%%%%%%%%%%%%%%%%%%%%%%%%%%%
%%%%%%%%%%%%%%%%%%%%%%%%%%%
%%%%%%%%%%%%%%%%%%%%%%%%%%%

\noindent
\textbf{Model Architecture.}
Figure \ref{fig:ops} top shows the proposed network to output conditional coordinate distributions. 
The top, middle and bottom branches model  $p(z_i|\textbf{s}_{\leq i-1})$, $p(y_i|\textbf{s}_{\leq i-1}, z_i)$ and  $p(x_i|\textbf{s}_{\leq i-1}, z_i, y_i)$, respectively.
% The point coordinates are sampled according to the estimated softmax probability distributions.
Note that the input points in the latter two cases are masked so that the network cannot see information not-yet-generated.
During training, points are available to compute all the context features, thus coordinate distributions can be estimated in parallel.
During the generative phase, the point coordinates are sampled according to the estimated softmax probability distributions.
This occurs sequentially, since each sampled coordinate needs to be fed as input back into the network, as shown in Figure \ref{fig:prog}.
Our proposed autoregressive architecture models point coordinate distribution categorically, thus a simple cross-entropy loss is sufficient to handle the shape learning process efficiently without requiring the use of a computationally-intensive set-to-set distance loss function.

%%%%%%%%%%%%%%%%%%%%%%%%%%%%%%%%%%%%%%%%%%
\noindent
\textbf{Conditional PointGrow.}
Given a condition or embedding vector, $\textbf{h}$, we intend to generate a shape satisfying the latent meaning of $\textbf{h}$.
To achieve this, Eq. (\ref{eq:1}) and (\ref{eq:2}) are adapted to Eq. (\ref{eq:3}) and (\ref{eq:4}), respectively, as below:
\begin{equation}
    p(\textbf{S}) = \prod_{i=1}^{n} p(\textbf{s}_i|\textbf{s}_1,...,\textbf{s}_{i-1}, \textbf{h}) = \prod_{i=1}^{n} p(\textbf{s}_i|\textbf{s}_{\leq i-1}, \textbf{h})
\label{eq:3}
\end{equation}
\vspace{-19pt}
\begin{multline}
    p(\textbf{s}_i|\textbf{s}_{\leq i-1}, \textbf{h}) = p(z_i|\textbf{s}_{\leq i-1}, \textbf{h}) \cdot p(y_i|\textbf{s}_{\leq i-1}, z_i, \textbf{h}) \\  \cdot p(x_i|\textbf{s}_{\leq i-1}, z_i, y_i, \textbf{h})
\label{eq:4}
\end{multline}
% \vspace{-1pt}
The additional condition, $\textbf{h}$, affects the coordinate distributions by adding biases and potential constrains in the generative process.
We implement this by changing the operation between adjacent fully-connected layers from $\textbf{x}^{i+1} = f(\textbf{W} \textbf{x}^i)$ to $\textbf{x}^{i+1} = f(\textbf{W} \textbf{x}^i + \textbf{H} \textbf{h})$, where $\textbf{x}^{i+1}$ and $\textbf{x}^{i}$ are feature vectors in the $i+1^{th}$ and $i^{th}$ layer, respectively, $\textbf{W}$ is a weight matrix, $\textbf{H}$ is a matrix that transforms $\textbf{h}$ into a vector with the same dimension as $\textbf{W} \textbf{x}^i$, and $f(\cdot)$ is a nonlinear activation function. 
In this paper, we experimented with $\textbf{h}$ as an one-hot categorical vector which adds class dependent bias, and an high-dimensional embedding vector of a 2D image which imposes geometric constraints.

%% file: sections/exp.tex
\section{Experiments}

%%%%%%%%%%%%%%%%%%%%%%%%%%%%%%
\noindent
\textbf{Datasets.}
We evaluated our framework on the ShapeNet \cite{chang2015shapenet} CAD dataset. We used a subset consisting of 17,687 models across 7 categories: airplanes, cars, tables, chairs, benches, cabinets and lamps. 
To generate corresponding point clouds, we sampled 10,000 points uniformly from each mesh file, and then used \textit{farthest point sampling} to select 1,024 points representing the shape.
Each category follows a split ratio 0.9/0.1 to separate training from testing sets. 
ModelNet40 \cite{wu20153d} and PASCAL3D+ \cite{xiang2014beyond} are used for additional analysis and demonstration. 
% ModelNet40 contains CAD models from 40 categories, and we obtain their point clouds from \cite{qi2017pointnet}. 
% PASCAL3D+ provides real images and is used to demonstrate the generalization ability of \emph{conditional PointGrow}. 

%%%%%%%%%%%%%%%%%%%%%%%%%%%%%%%%%%%%%%%%%%
\begin{figure*}[t!]
\vspace{-15pt}
\begin{center}
\includegraphics[width=0.9\linewidth]{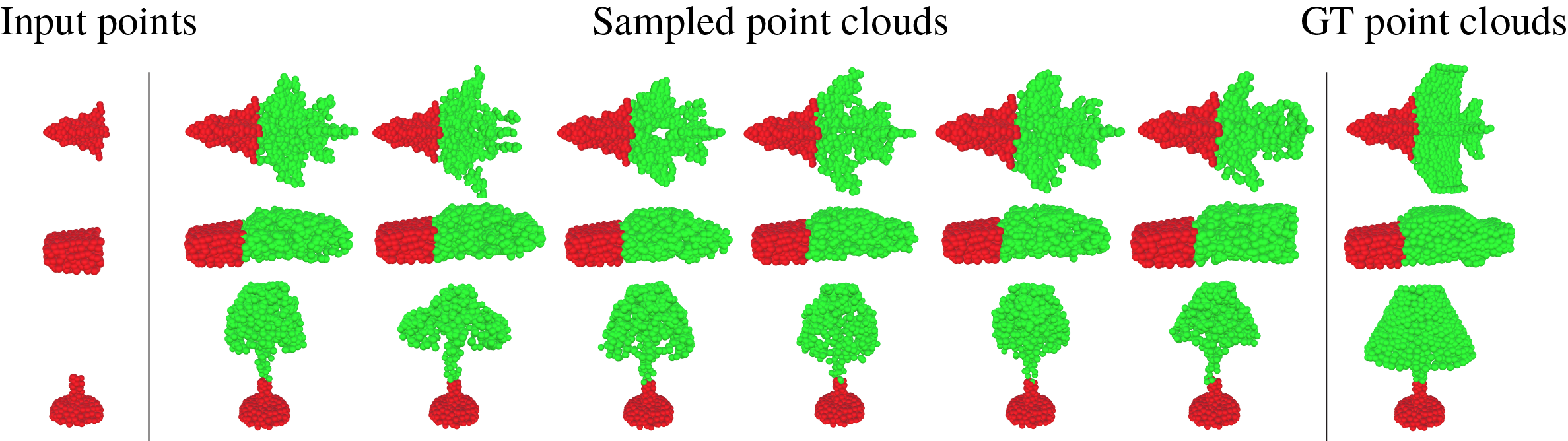}
\end{center}
\vspace{-10pt}
\caption{Shape completion results generated by \emph{PointGrow}.}
\label{fig:shape_compl}
\vspace{-7pt}
\end{figure*}
%%%%%%%%%%%%%%%%%%%%%%%%%%%%%%%%%%%%%%%%%%

% %%%%%%%%%%%%%%%%%%%%%%%%%%%%%%%%
\subsection{Unconditional point cloud generation}
We first evaluate \emph{unconditional PointGrow} by addressing the following questions:
(1) Can the model generate visually plausible 3D shapes in an interpretable manner? 
(2) Can the model generate diverse shapes?
(3) Is the proposed self-attention module important for shape generation?
(4) Does the model learn meaningful feature representation? 

% %%%%%%%%%%%%%%%%%%%%%%%%%%%%%%%%
\noindent
\textbf{(1) Generative Process Interpretability and Shape Quality.}
We start evaluation by showing qualitative results of generated point clouds of \emph{unconditional PointGrow}, shown in Figure \ref{fig:attention_map}.
Fine-detailed structures can be observed from the generated shapes, such as the jet engine of airplanes and legs of furniture (e.g. table and chair).
In the bottom part of Figure \ref{fig:attention_map}, we additionally show attention fields when generating corresponding query points in the process.
It can be observed that the model focuses on different regions when generating points for different parts, and the focused areas are usually semantically related no matter their spatial distances.
For example, when generating airplane wing points, the model aggregates structural knowledge from available wing part points; when generating points close to table legs, other previously generated leg points contribute most; when generating torchiere shade points for the lamp, the model considers both existing torchiere shade and base areas for a proper structural match.
We also show a sampled generative process for an airplane in Figure \ref{fig:prog}.
Note that the model produces different conditional distribution for points at different parts (e.g. in the second row of Figure \ref{fig:prog}, the network model outputs a roughly symmetric distribution along the $X$ axis, describing the airplane's wings). 
From the above investigation, the interpretability of the proposed generative process is demonstrated.
In this experiment, we train our model for each category separately, because an unconditional model lacks knowledge about the target shape when sampling from scratch. In later experiments, we show that when a categorical condition is given, it is possible to train the model across multiple shape categories and generate plausible shapes.

\begin{figure}[h!]
\vspace{-8pt}
\centering
\includegraphics[width=1.0\linewidth]{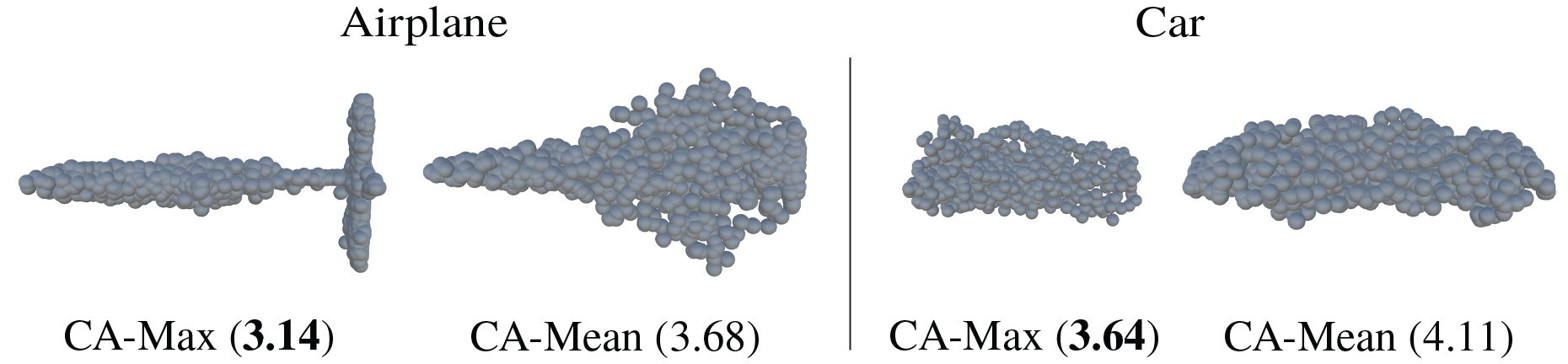}
\caption{Negative log-likelihood for CA-Max and CA-Mean baselines on ShapeNet airplane and car testing sets.}
\label{fig:mean_max}
\vspace{-10pt}
\end{figure}

Next, we quantitatively evaluate the quality of generated shapes.
The negative log-likelihood is commonly used to evaluate autoregressive models for image and audio generation \cite{oord2016pixel, van2016wavenet}.
However, we observed inconsistency between this value and the visual quality of generated 3D shapes. 
This is validated by comparing two baseline models: \textit{CA-Mean} and \textit{CA-Max}, where the SACA operation is replaced with the CA operation implemented by mean and max pooling, respectively.
In Figure \ref{fig:mean_max}, we report negative log-likelihoods in {\it bits per coordinate} on ShapeNet testing sets of airplane and car categories, and visualize their representative results. 
Despite CA-Max shows lower negative log-likelihood values, it gives less visually plausible results (i.e. airplanes lose wings and cars lose rear ends).

% classificaiton.
To actually evaluate the generated shape quality, we argue that if generated 3D shapes contain consistent semantic features as real shapes, a classification model trained on real shapes should perform well on generated ones, and vice versa.
Therefore, after training on ShapeNet sets, we generate 300 point clouds per category (2,100 in total for 7 categories), and conduct two classification tasks: one training on original ShapeNet training sets and testing on generated shapes, the other training on generated shapes and testing on original ShapeNet testing sets.
Here, PointNet \cite{qi2017pointnet}, a widely-uesd model, is used as the point cloud shape classifier. 
We implement another two GAN-based competing methods and report classification results in Table \ref{table:cls}, together with model complexity using number of model parameters.
In the first classification task, our SACA-A model outperforms existing models by a relatively large margin, while in the second task, SACA-A and SACA-B models show similar performance.

\begin{table}[h!]
\footnotesize
\vspace{-10pt}
\begin{center}
\begin{tabular}{lccc}
\multicolumn{1}{l}{Methods}  & 
\multicolumn{1}{c}{SG} & 
\multicolumn{1}{c}{GS} &
\multicolumn{1}{c}{\# parameters}
\\ \hline
3D-GAN \cite{wu2016learning} & 82.7 & 83.4 & 22.53M\\
Latent-GAN \cite{achlioptas2017representation} & 81.6 & 82.7 & 15.78M \\
Baseline (CA-Max) & 71.9 & 83.4 & 0.23M \\
Baseline (CA-Mean) & 82.1 & 84.4 & 0.23M \\
Ours (SACA-A) & \textbf{90.3} & 91.8 & 0.29M\\
Ours (SACA-B) & 89.4 & \textbf{91.9} & 0.25M\\
\hline 
\end{tabular}
\end{center}
\vspace{-5pt}
\caption{Classification accuracy using PointNet \cite{qi2017pointnet}. \textbf{SG:} Training on ShapeNet sets and testing on generated shapes; \textbf{GS:} Training on generated shapes and testing on ShapeNet sets.}
\vspace{-10pt}
\label{table:cls}
\end{table}

%%%%%%%%%%%%%%%%%%%%%%%%%%%%%%%%%%%%%%%%%%
\noindent
\textbf{(2) Shape Diversity.}
To demonstrate \emph{PointGrow} can generate diverse shapes, we conducted a shape completion task.
Given an initial set of points, our model is capable of completing shapes in multiple ways. 
Figure \ref{fig:shape_compl} visualizes examples.
The input points are sampled from ShapeNet testing sets, which are not seen during the training process.
The shapes generated by our model are different from the original ground truth point clouds, but still look plausible.
A current limitation of our model is that it works only when the input point set is given as the beginning part of a shape along its primary axis, and in future work we will investigate how to complete shapes when partial point clouds are given from any directions.

% %user study
% To overall access the quality and diversity of generated shapes, we also conduct a user study evaluation \wrt two aspects, fidelity and diversity.
% Here, we consider CA-Max, CA-Mean, Latent-Gan \cite{achlioptas2017representation} as comparing methods.
% We randomly select 10 generated airplane and car point clouds from each method. To calculate the fidelity score, we ask the user to score $0$, $0.5$ or $1.0$ for each shape, and take the average of them. The diversity score is obtained by asking the user to scale from $0.1$ to $1.0$ with an interval of $0.1$ about the generated shape diversity within each method. 10 subjects without computer vision background participated in this test. The user study results are plotted in Figure \ref{fig:user_study}. We observe that (1) CA-Mean is more favored than CA-Max, and (2) our \emph{PointGrow} receives the highest preference on both fidelity and diversity. 

% % % figure
% \begin{figure}[h!]
% \vspace{-6pt}
% \centering
% \includegraphics[width=1.0\linewidth]{./sections/figures/userstudy-3.png}
% \vspace{-15pt}
% \caption{User study results on fidelity and diversity of the generated point clouds.}
% \label{fig:user_study}
% \vspace{-10pt}
% \end{figure}

%%%%%%%%%%%%%%%%%%%%%%%%%%%%%%%%%%%%%%%%%%
\noindent
\textbf{(3) Ablation Study on Self-Attention Module.}
We conduct an ablation study to investigate the importance of the Self-Attention module in shape generation process.
% We found that without CA modules, it is hard to generate visually-plausible shapes.
To quantitatively evaluate generated shapes and inspired by Fr\'echet Inception Distance (FID) \cite{heusel2017gans, zhang2018self}, a metric widely used to measure the visual quality of generated images, we provide a similar metric, ``PointNet Distance" (PND), to measure the geometric quality of generated point clouds.
PND follows the same assumption and formula as FID, except that the features from a point cloud is obtained from the global feature of a PointNet model pre-trained on ModelNet40 with a global feature dimension of 128.
Three model architectures are considered: without the second CA module (No CA), CA-Mean baseline, and SACA-A.
We generate 200 point clouds per category for each model architecture.
% In FID, the number of samples used to calculate Gaussian statistics should be greater than the dimension of feature vectors, so here we change the global feature dimension of PointNet from 1024 to 128 during the pre-training process.
The PND is computed for each category and reported in Table \ref{table:ablation}. 
We observe a large performance improvement by gathering context information with self-attention supported.

%%%%%%%%%%%%%%%%%%%%%%%%%%%%%%%%%%%%%%%%%%
%%%%%%%%%%%%%%%%%%%%%%%%%%%%%%%%%%%%%%%%%%
\begin{table}[h!]
\vspace{-10pt}
%\footnotesize
\begin{center}
\scalebox{0.65}{
\begin{tabular}{lcccccccc}
\multicolumn{1}{l}{}  & 
\multicolumn{1}{l}{airplane}  & 
\multicolumn{1}{c}{car} & 
\multicolumn{1}{c}{table} & 
\multicolumn{1}{c}{chair} & 
\multicolumn{1}{c}{bench} &
\multicolumn{1}{c}{cabinet} &
\multicolumn{1}{c}{lamp} &
\multicolumn{1}{c}{Avg.} 
\\ \hline
No CA   & 54.08 & 58.30 & 219.54 & 181.18 & 104.38 & 318.89 & 165.52 & 157.41 \\
CA-Mean & 7.05 & 4.33 & 12.83 & 27.47 & 6.09 & 20.67 & 88.14 & 23.79 \\
SACA-A  &  {\bf 1.92} & {\bf 0.53} & {\bf 2.40} & {\bf  4.61} & {\bf 1.24} & {\bf 3.67} & {\bf 8.34} & {\bf 3.24} \\
% SACA-B  & 1.94 & {\bf 0.38} & 2.51 & {\bf 3.82} & {\bf 1.10} & {\bf 3.63} & {\bf 2.83} & {\bf 2.32} \\
\hline 
\end{tabular}
}
\end{center}
\caption{The PointNet Distance (PND) for each category. A lower score signifies a better model.}
\label{table:ablation}
\end{table}
%%%%%%%%%%%%%%%%%%%%%%%%%%%%%%%%%%%%%%%%%%
%%%%%%%%%%%%%%%%%%%%%%%%%%%%%%%%%%%%%%%%%%

%%%%%%%%%%%%%%%%%%%%%%%%%%%%%%%%%%%%%%%%
% Table
%%%%%%%%%%%%%%%%%%%%%%%%%%%%%%%%%%%%%%%%
\begin{table}[h!]
\vspace{-10pt}
\footnotesize
\vspace{-10pt}
\begin{center}
\begin{tabular}{lcc}
\multicolumn{1}{l}{Methods}  & 
\multicolumn{1}{c}{SVM} &
\multicolumn{1}{c}{SL}
\\ \hline
SPH \cite{kazhdan2003rotation} & 68.2 & - \\
LFD \cite{chen2003visual}      & 75.2 & - \\
T-L Network \cite{girdhar2016learning} & 74.4 & - \\
VConv-DAE \cite{sharma2016vconv} & 75.5 & -\\
3D-GAN \cite{wu2016learning} & 83.3 & -\\
Latent-GAN-EMD \cite{achlioptas2017representation} & 84.0 & - \\
Latent-GAN-CD \cite{achlioptas2017representation} & 84.5 & - \\
MTN \cite{gadelha2018multiresolution} & - & \textbf{86.4} \\
Ours (SACA-A) & \textbf{85.8} & 86.3 \\
Ours (SACA-B) & 84.4 & 85.3 \\
\hline 
\end{tabular}
\end{center}
\vspace{-5pt}
\caption{The comparison on classification accuracy between our models and other unsupervised methods on ModelNet40 dataset using linear SVM classifier and single layer classifier (SL). }
\label{table:cls_unsupervised}
\vspace{-10pt}
\end{table}
%%%%%%%%%%%%%%%%%%%%%%%%%%%%%%%%%%%%%%%%
%%%%%%%%%%%%%%%%%%%%%%%%%%%%%%%%%%%%%%%%

%%%%%%%%%%%%%%%%%%%%%%%%%%%%%%%%%%%%%%%%%%
\noindent
\textbf{(4) Unsupervised Feature Learning.}
To prove the model actually learns meaningful representations, we extract learned features and use them for classification tasks. 
We obtain the feature vector of a shape by applying different types of ``symmetric" functions as illustrated in \cite{qi2017pointnet} (i.e. min, max and mean pooling) on features of each layer before the SACA operation, and concatenate them all. 
Following \cite{wu2016learning}, we pre-train our model on 7 categories from the ShapeNet dataset, and then use this model to extract feature vectors for both training and testing shapes from the ModelNet40 dataset.
We experimented with both linear SVM and single layer classifiers, following the same settings as \cite{wu2016learning} and \cite{gadelha2018multiresolution}, respectively.
% During the pre-training phase, random rotations are applied to the input point clouds along the gravity axis.
We report our best results in Table \ref{table:cls_unsupervised}.
The SACA-A model achieves the best performance using SVM classifier, and performs slightly worse than MTN \cite{gadelha2018multiresolution} using single layer classifier.

%%%%%%%%%%%%%%%%%%%%%%%%%%%%%%%%%%%%%%%%%%
%%%%%%%%%%%%%%%%%%%%%%%%%%%%%%%%%%%%%%%%%%
\subsection{Conditional Point Cloud Generation}
% SACA-A is used to demonstrate \emph{conditional PointGrow} in this subsection.
To evaluate \emph{conditional PointGrow}, we answer the following questions:
(1) Can the model be trained across multiple categories?
(2) Can the model generate plausible shapes for image conditions?

\begin{figure}[h!]
\vspace{-7pt}
\begin{center}
\includegraphics[width=1.0\linewidth]{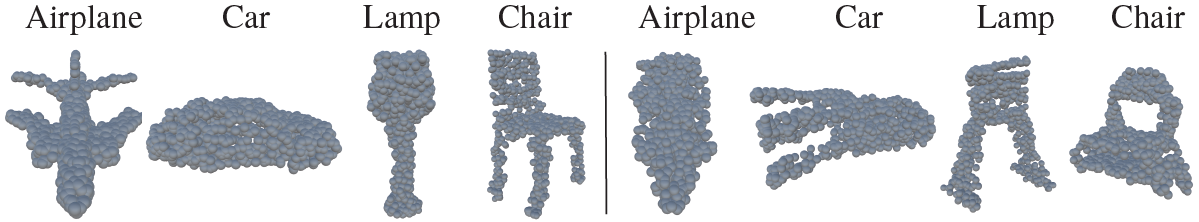}
\end{center}
\vspace{-8pt}
\caption{Generated point clouds of \emph{PointGrow} conditioned on different one-hot categorical labels. Right part shows failure cases.}
\vspace{-8pt}
\label{fig:one_hot}
\end{figure}

%%%%%%%%%%%%%%%%%%%%%%%%%%%%%%%%%%%%%%%%%%
\noindent
\textbf{(1) Conditioned on Category Label.}
We first experiment with category-conditional modelling of point clouds, given an one-hot vector $\textbf{h}$ with its nonzero element $h_{i}$ indicating the $i^{th}$ shape category.
The one-hot condition provides categorical knowledge to guide the shape generation process, enabling the model to be trained across multiple categories.
Figure \ref{fig:one_hot} shows generated shape examples.
Failure cases are also observed: generated shapes present interwoven geometric properties from other shape types.  
For example, the airplane misses wings and generates a car-like body; the lamp and the car develop chair leg structures.

\begin{figure*}[h!]
\vspace{-28pt}
\begin{center}
\includegraphics[width=1.0\linewidth]{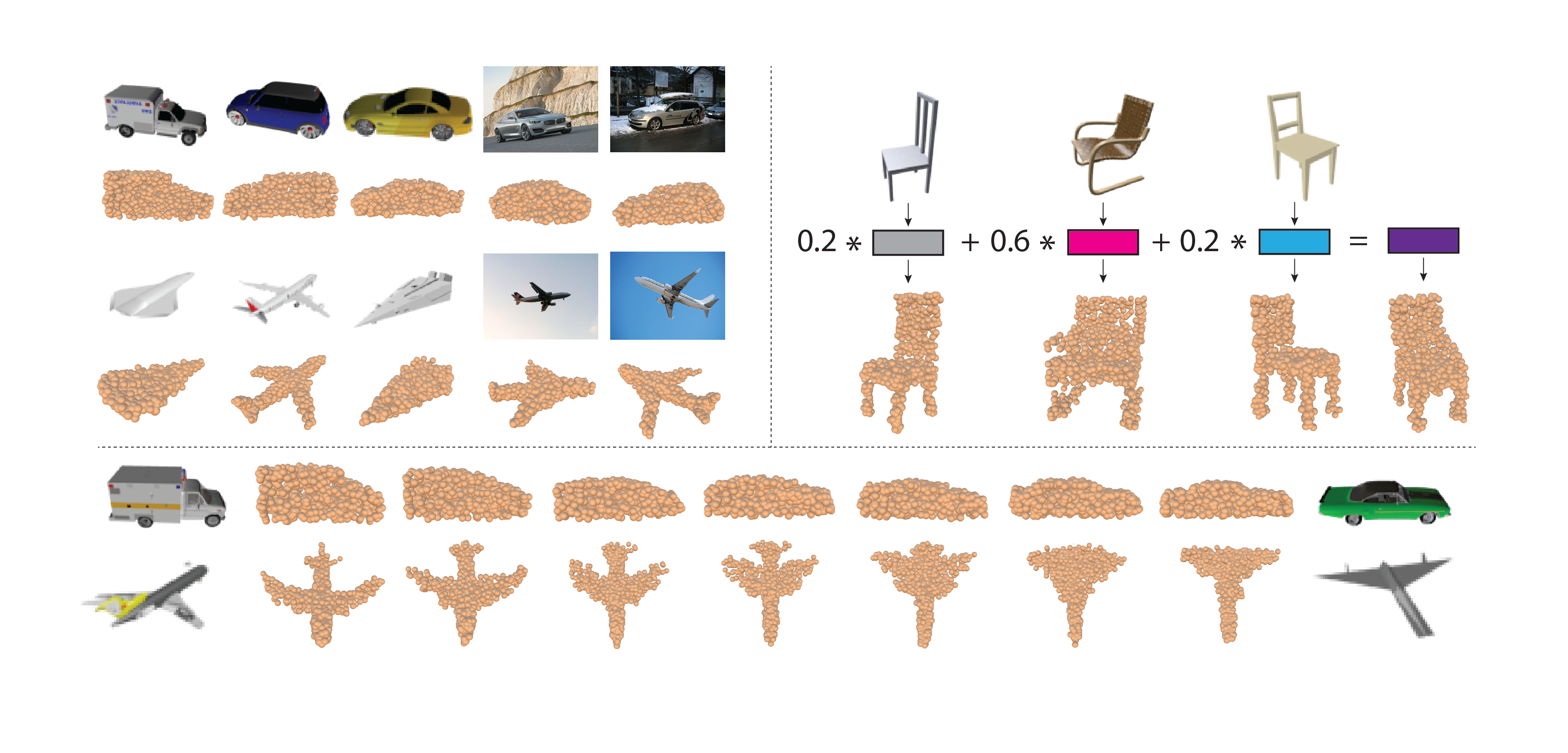}
\end{center}
\vspace{-33pt}
\caption{\textbf{Upper left:} Generated point clouds conditioned on synthetic testing images from ShapeNet (first 4 columns) and real images from PASCAL3D+ (last 2 columns). \textbf{Upper right:} Examples of image condition arithmetic for chairs. \textbf{Bottom:} Examples of image condition linear interpolation for cars. Condition vectors from leftmost and rightmost images are used as endpoints for the shape interpolation.}
% \vspace{-5pt}
\label{fig:im_cond}
\end{figure*}

\begin{table*}[h!]
\vspace{-8pt}
\small
\begin{center}
\setlength\tabcolsep{2.0pt}
\begin{tabular}{lcccccccccccccc}
\multicolumn{1}{l}{}  & 
\multicolumn{1}{c}{airplane} & 
\multicolumn{1}{c}{bench} &
\multicolumn{1}{c}{cabinet} &
\multicolumn{1}{c}{car} &
\multicolumn{1}{c}{chair} &
\multicolumn{1}{c}{monitor} &
\multicolumn{1}{c}{lamp} &
\multicolumn{1}{c}{speaker} &
\multicolumn{1}{c}{firearm} &
\multicolumn{1}{c}{couch} &
\multicolumn{1}{c}{table} &
\multicolumn{1}{c}{cellphone} &
\multicolumn{1}{c}{watercraft} &
\multicolumn{1}{c}{Avg.}
% {} & airplane & bench & cabinet & car & chair & monitor & lamp & speaker & firearm & couch & table & cellphone & watercraft & avg.
\\ \hline
3D-R2N2 (1 view) & 0.513 & 0.421 & 0.716 & 0.798 & 0.466 & 0.468 & 0.381 & 0.662 & 0.544 & 0.628 & 0.513 & 0.661 & 0.513 & 0.560 \\
3D-R2N2 (5 views) & 0.561 & 0.527 & {\bf 0.772} & 0.836 & {\bf 0.550} & 0.565 & 0.421 & 0.717 & 0.600 & 0.706 & 0.580 & {\bf 0.754} & 0.610 & 0.631 \\
PointSetGen & 0.601 & 0.550 & 0.771 & 0.831 & 0.544 & 0.552 & 0.462 & {\bf 0.737} & 0.604 & {\bf 0.708} & {\bf 0.606} & 0.749 & 0.611 & 0.640 \\
Ours & {\bf 0.742} & {\bf 0.629} & 0.675 & {\bf 0.839} & 0.537 & {\bf 0.567} & {\bf 0.560} & 0.569 & {\bf 0.674} & 0.676 & 0.590 & 0.729 & {\bf 0.737} & {\bf 0.656} \\
\hline 
\end{tabular}
\end{center}
\vspace{-6pt}
\caption{Conditional generation evaluation by per-category IoU on 13 ShapeNet categories. We compare our results with 3D-R2N2 (\cite{choy20163d}) and PointSetGen \cite{fan2017point}.}
\label{table:iou}
\end{table*}

\begin{table*}[h!]
\vspace{-10pt}
\small
\begin{center}
\setlength\tabcolsep{2.0pt}
\begin{tabular}{lcccccccccccccc}
\multicolumn{1}{l}{}  & 
\multicolumn{1}{c}{airplane} & 
\multicolumn{1}{c}{bench} &
\multicolumn{1}{c}{cabinet} &
\multicolumn{1}{c}{car} &
\multicolumn{1}{c}{chair} &
\multicolumn{1}{c}{monitor} &
\multicolumn{1}{c}{lamp} &
\multicolumn{1}{c}{speaker} &
\multicolumn{1}{c}{firearm} &
\multicolumn{1}{c}{couch} &
\multicolumn{1}{c}{table} &
\multicolumn{1}{c}{cellphone} &
\multicolumn{1}{c}{watercraft} &
\multicolumn{1}{c}{Avg.}
% {} & airplane & bench & cabinet & car & chair & monitor & lamp & speaker & firearm & couch & table & cellphone & watercraft & avg.
\\ \hline
3D-R2N2 (1 view) & 3.207 & 3.350 & 1.636 & 1.808 & 2.759 & 3.235 & 8.400 & 2.652 & 4.798 & 2.725 & 3.118 & 2.202 & 3.592 & 3.345 \\
3D-R2N2 (5 views) & 2.399 & 2.323 & 1.420 & 1.664 & 1.854 & 2.088 & 5.698 & 2.487 & 4.193 & 2.306 & 2.128 & 1.874 & 3.210 & 2.588 \\
PointSetGen (1 view) & 1.301 & 1.814 & 2.463 & 1.800 & 1.887 & 1.919 & 2.347 & 3.215 & 1.316 & 2.592 & 1.874 & 1.516 & 1.715 & 1.982 \\
Lin et al. (1 view) & 1.294 & 1.757 & 1.814 & 1.446 & 1.886 & 2.142 & 2.635 & 2.371 & 1.289 & 1.917 & 1.689 & 1.939 & 1.813 & 1.846 \\ 
MRTNet (1 view) & 0.976 & {\bf 1.438} & 1.774 & 1.395 & {\bf 1.650} & {\bf 1.815} & {\bf 1.944} & 2.165 & {\bf 1.029} & 1.768 & {\bf 1.570} & 1.346 & 1.394 & 1.559 \\ 
Ours (1 view) & {\bf 0.615} & 1.726 & {\bf 1.201} & {\bf 0.416} & 1.775 & 1.937 & 2.235 & {\bf 1.998} & 1.300 & {\bf 1.405} & 1.974 & {\bf 0.765} & {\bf 0.865} & {\bf 1.401} \\
\hline 
3D-R2N2 (1 view) & 2.879 & 3.697 & 2.817 & 3.238 & 4.207 & 4.283 & 9.722 & 4.335 & 2.996 & 3.628 & 4.208 & 3.314 & 4.007 & 4.102 \\
3D-R2N2 (5 views) & 2.391 & 2.603 & 2.619 & 3.146 & 3.080 & 2.953 & 7.331 & 4.203 & 2.447 & 3.196 & 3.134 & 2.734 & 3.614 & 3.342 \\
PointSetGen (1 view) & 1.488 & 1.983 & 2.444 & 2.053 & 2.355 & 2.334 & 2.212 & 2.788 & 1.358 & 2.784 & 2.229 & 1.989 & 1.877 & 2.146 \\
Lin et al. (1 view) & 1.541 & 1.487 & {\bf 1.072} & 1.061 & 2.041 & {\bf 1.440} & 4.459 & {\bf 1.706} & 1.510 & 1.423 & 1.620 & 1.198 & 1.550 & 1.701 \\
MRTNet (1 view) & 0.920 & {\bf 1.326} & 1.602 & 1.303 & {\bf 1.603} & 1.901 & 2.089 & 2.121 & {\bf 1.028} & 1.756 & {\bf 1.570} & 1.332 & 1.490 & 1.529 \\
Ours (1 view) & {\bf 0.914} & 1.531 & 1.564 & {\bf 0.732} & 1.857 & 1.789 & {\bf 1.732} & 2.020 & 1.166 & {\bf 1.381} & 1.744 & {\bf 0.869} & {\bf 1.070} & {\bf 1.413} \\
\hline 
\end{tabular}
\end{center}
\vspace{-7pt}
\caption{Single-image shape inference results. Top part: pred $\rightarrow$ GT errors; Bottom part: GT $\rightarrow$ pred errors, both scaled by 100. We compare our results with 3D-R2N2 \cite{choy20163d}, PointSetGen \cite{fan2017point}, Lin et al. \cite{lin2017learning} and MRTNet \cite{gadelha2018multiresolution}.}
\vspace{-10pt}
\label{table:dis_err}
\end{table*} 

%%%%%%%%%%%%%%%%%%%%%%%%%%%%%%%%%%%%%%%%%%
\noindent
\textbf{(2) Conditioned on 2D Image.}
Next, we experiment with image conditions for point cloud generation.
Image conditions add additional constrains to the point cloud generation process such that the geometric structures of sampled shapes match their 2D projections.
In our experiments, we obtain an image condition vector through an image encoder, and optimize it together with the rest model components from scratch. 
The model is trained on synthetic ShapeNet dataset, and one out of 24 views of a shape (provided by \cite{choy20163d}) is selected as the image condition input.
The trained model is also tested on foreground objects of real images from the PASCAL3D+ dataset to prove its generalizability. 
% For each real image input, we cropped and centralized foreground objects.
The PASCAL3D+ dataset is challenging because the images are captured in real environments.
% and contain noisier visual signals which are not seen during the training process.
% We show ShapeNet testing images and PASCAL3D+ real images together with their sampled point cloud results on Figure \ref{fig:im_cond} upper left.
Testing examples are shown on Figure \ref{fig:im_cond} upper left.

We quantitatively evaluate the conditional generation results in terms of mean Intersection-over-Union (mIoU) \cite{choy20163d} and point-wise 3D Euclidean distance \cite{lin2017learning}.
Here we obtain ground truth points as voxel centers of 3D volumes from \cite{choy20163d}, and only consider shapes containing more than 500 occupied voxels, with 500 of them uniformly sampled to describe the shape.
To compensate for the sampling randomness of \emph{PointGrow}, we align generated points to their nearest voxels within a neighborhood of 2-voxel radius.
When calculating surface-to-surface 3D Euclidean distance metric, we further remove interior points with 26 non-empty neighbors for fair comparison when selecting ground truth points.
As shown in Table~\ref{table:iou} and Table \ref{table:dis_err}, \emph{PointGrow} achieves above-par performance on conditional 3D shape generation.

Further, we demonstrate that intermediate 3D shapes can be generated from linearly interpolated embedding vectors of image pairs (Figure \ref{fig:im_cond} bottom), and compositive shapes can be generated by applying arithmetic on embedded image condition vectors (Figure \ref{fig:im_cond} upper right).

%% file: sections/conclusion.tex
\section{Discussion and Conclusion}
This work studies the problem of point cloud generation. Unlike previous work, which minimizes set-to-set distances for generative learning, our model builds upon an autoregressive architecture and exploits point-to-point relations during generation, allowing the generative process to be better understood and interpreted.
To further capture long-range dependencies in a self-adaptive manner and address the irregularity of point cloud data, two self-attention models are integrated within our framework. 
Extensive experiments validates the efficacy of this approach across a wide range of tasks. 

Though our model generates visually-plausible 3D shapes, it faces two potential limitations. Firstly, due to the iterative property intrinsic to autoregressive models, the model scales poorly when generating large point sets. Recent work \cite{ramachandran2017fast, oord2017parallel} has accelerated autoregression-based audio generation.
Similar techniques are also applicable here (\eg generating clouds in a hierarchical rather than sequential manner).
% since clouds can be stored using KD trees \cite{brown2014building} for faster searching). 
Secondly, our model only generates a point cloud along its primary axis as determined in training. This does not hinder generation performance if the shape is sampled from scratch, but limits its applicability to applications like shape completion. Generating clouds more flexibly will be an important topic for further research.

% In summary, this manuscript has identified the importance of interpoint dependencies for point cloud generation, and we hope this work serves as a basis for additional exploration including autoregressive models and other models for interpreting, understanding, and exploiting \emph{internal relations} for 3D generative learning. 

%% file: main.bbl
\begin{thebibliography}{1}\itemsep=-1pt

\bibitem{Alpher02}
A.~Alpher.
\newblock Frobnication.
\newblock {\em Journal of Foo}, 12(1):234--778, 2002.

\bibitem{Alpher03}
A.~Alpher and J.~P.~N. Fotheringham-Smythe.
\newblock Frobnication revisited.
\newblock {\em Journal of Foo}, 13(1):234--778, 2003.

\bibitem{Alpher04}
A.~Alpher, J.~P.~N. Fotheringham-Smythe, and G.~Gamow.
\newblock Can a machine frobnicate?
\newblock {\em Journal of Foo}, 14(1):234--778, 2004.

\bibitem{Authors06b}
Authors.
\newblock Frobnication tutorial, 2006.
\newblock Supplied as additional material {\tt tr.pdf}.

\bibitem{Authors06}
Authors.
\newblock The frobnicatable foo filter, 2011.
\newblock Face and Gesture submission ID 324. Supplied as additional material
  {\tt fg324.pdf}.

\end{thebibliography}


\begin{thebibliography}{10}\itemsep=-1pt

\bibitem{achlioptas2017representation}
P.~Achlioptas, O.~Diamanti, I.~Mitliagkas, and L.~Guibas.
\newblock Representation learning and adversarial generation of 3d point
  clouds.
\newblock {\em arXiv preprint arXiv:1707.02392}, 2017.

\bibitem{achlioptas2018learning}
P.~Achlioptas, O.~Diamanti, I.~Mitliagkas, and L.~Guibas.
\newblock Learning representations and generative models for 3d point clouds.
\newblock 2018.

\bibitem{bahdanau2014neural}
D.~Bahdanau, K.~Cho, and Y.~Bengio.
\newblock Neural machine translation by jointly learning to align and
  translate.
\newblock {\em arXiv preprint arXiv:1409.0473}, 2014.

\bibitem{boscaini2016learning}
D.~Boscaini, J.~Masci, E.~Rodol{\`a}, and M.~Bronstein.
\newblock Learning shape correspondence with anisotropic convolutional neural
  networks.
\newblock In {\em Advances in Neural Information Processing Systems}, pages
  3189--3197, 2016.

\bibitem{brock2016generative}
A.~Brock, T.~Lim, J.~M. Ritchie, and N.~Weston.
\newblock Generative and discriminative voxel modeling with convolutional
  neural networks.
\newblock {\em arXiv preprint arXiv:1608.04236}, 2016.

\bibitem{bruna2013spectral}
J.~Bruna, W.~Zaremba, A.~Szlam, and Y.~LeCun.
\newblock Spectral networks and locally connected networks on graphs.
\newblock {\em arXiv preprint arXiv:1312.6203}, 2013.

\bibitem{chang2015shapenet}
A.~X. Chang, T.~Funkhouser, L.~Guibas, P.~Hanrahan, Q.~Huang, Z.~Li,
  S.~Savarese, M.~Savva, S.~Song, H.~Su, et~al.
\newblock Shapenet: An information-rich 3d model repository.
\newblock {\em arXiv preprint arXiv:1512.03012}, 2015.

\bibitem{chen2003visual}
D.-Y. Chen, X.-P. Tian, Y.-T. Shen, and M.~Ouhyoung.
\newblock On visual similarity based 3d model retrieval.
\newblock In {\em Computer graphics forum}, 2003.

\bibitem{cheng2016long}
J.~Cheng, L.~Dong, and M.~Lapata.
\newblock Long short-term memory-networks for machine reading.
\newblock {\em arXiv preprint arXiv:1601.06733}, 2016.

\bibitem{choy20163d}
C.~B. Choy, D.~Xu, J.~Gwak, K.~Chen, and S.~Savarese.
\newblock 3d-r2n2: A unified approach for single and multi-view 3d object
  reconstruction.
\newblock In {\em ECCV}, 2016.

\bibitem{denil2012learning}
M.~Denil, L.~Bazzani, H.~Larochelle, and N.~de~Freitas.
\newblock Learning where to attend with deep architectures for image tracking.
\newblock {\em Neural computation}, 24(8):2151--2184, 2012.

\bibitem{fan2017point}
H.~Fan, H.~Su, and L.~J. Guibas.
\newblock A point set generation network for 3d object reconstruction from a
  single image.
\newblock In {\em CVPR}, volume~2, page~6, 2017.

\bibitem{gadelha2017shape}
M.~Gadelha, S.~Maji, and R.~Wang.
\newblock Shape generation using spatially partitioned point clouds.
\newblock {\em arXiv preprint arXiv:1707.06267}, 2017.

\bibitem{gadelha2018multiresolution}
M.~Gadelha, R.~Wang, and S.~Maji.
\newblock Multiresolution tree networks for 3d point cloud processing.
\newblock {\em arXiv preprint arXiv:1807.03520}, 2018.

\bibitem{girdhar2016learning}
R.~Girdhar, D.~F. Fouhey, M.~Rodriguez, and A.~Gupta.
\newblock Learning a predictable and generative vector representation for
  objects.
\newblock In {\em ECCV}, 2016.

\bibitem{graham2017submanifold}
B.~Graham and L.~van~der Maaten.
\newblock Submanifold sparse convolutional networks.
\newblock {\em arXiv preprint arXiv:1706.01307}, 2017.

\bibitem{groueix2018atlasnet}
T.~Groueix, M.~Fisher, V.~G. Kim, B.~C. Russell, and M.~Aubry.
\newblock Atlasnet: A papier-m$\backslash$\^{} ach$\backslash$'e approach to
  learning 3d surface generation.
\newblock {\em arXiv preprint arXiv:1802.05384}, 2018.

\bibitem{hane2017hierarchical}
C.~H{\"a}ne, S.~Tulsiani, and J.~Malik.
\newblock Hierarchical surface prediction for 3d object reconstruction.
\newblock In {\em 3D Vision (3DV), 2017 International Conference on}, pages
  412--420. IEEE, 2017.

\bibitem{heusel2017gans}
M.~Heusel, H.~Ramsauer, T.~Unterthiner, B.~Nessler, and S.~Hochreiter.
\newblock Gans trained by a two time-scale update rule converge to a local nash
  equilibrium.
\newblock In {\em Advances in Neural Information Processing Systems}, pages
  6626--6637, 2017.

\bibitem{jack2018learning}
D.~Jack, J.~K. Pontes, S.~Sridharan, C.~Fookes, S.~Shirazi, F.~Maire, and
  A.~Eriksson.
\newblock Learning free-form deformations for 3d object reconstruction.
\newblock {\em arXiv preprint arXiv:1803.10932}, 2018.

\bibitem{jiang2018gal}
L.~Jiang, S.~Shi, X.~Qi, and J.~Jia.
\newblock Gal: Geometric adversarial loss for single-view 3d-object
  reconstruction.
\newblock In {\em European Conference on Computer Vision}, pages 820--834.
  Springer, Cham, 2018.

\bibitem{kanazawa2018learning}
A.~Kanazawa, S.~Tulsiani, A.~A. Efros, and J.~Malik.
\newblock Learning category-specific mesh reconstruction from image
  collections.
\newblock {\em arXiv preprint arXiv:1803.07549}, 2018.

\bibitem{kazhdan2003rotation}
M.~Kazhdan, T.~Funkhouser, and S.~Rusinkiewicz.
\newblock Rotation invariant spherical harmonic representation of 3 d shape
  descriptors.
\newblock In {\em Symposium on geometry processing}, 2003.

\bibitem{klokov2017escape}
R.~Klokov and V.~Lempitsky.
\newblock Escape from cells: Deep kd-networks for the recognition of 3d point
  cloud models.
\newblock In {\em ICCV}, 2017.

\bibitem{kong2017using}
C.~Kong, C.-H. Lin, and S.~Lucey.
\newblock Using locally corresponding cad models for dense 3d reconstructions
  from a single image.
\newblock In {\em Proceedings of the IEEE Conference on Computer Vision and
  Pattern Recognition}, volume~2, 2017.

\bibitem{larochelle2010learning}
H.~Larochelle and G.~E. Hinton.
\newblock Learning to combine foveal glimpses with a third-order boltzmann
  machine.
\newblock In {\em Advances in neural information processing systems}, pages
  1243--1251, 2010.

\bibitem{li2018pointcnn}
Y.~Li, R.~Bu, M.~Sun, and B.~Chen.
\newblock Pointcnn.
\newblock {\em arXiv preprint arXiv:1801.07791}, 2018.

\bibitem{lin2017learning}
C.-H. Lin, C.~Kong, and S.~Lucey.
\newblock Learning efficient point cloud generation for dense 3d object
  reconstruction.
\newblock {\em arXiv preprint arXiv:1706.07036}, 2017.

\bibitem{masci2015geodesic}
J.~Masci, D.~Boscaini, M.~Bronstein, and P.~Vandergheynst.
\newblock Geodesic convolutional neural networks on riemannian manifolds.
\newblock In {\em Proceedings of the IEEE international conference on computer
  vision workshops}, pages 37--45, 2015.

\bibitem{maturana2015voxnet}
D.~Maturana and S.~Scherer.
\newblock Voxnet: A 3d convolutional neural network for real-time object
  recognition.
\newblock In {\em Intelligent Robots and Systems (IROS), 2015 IEEE/RSJ
  International Conference on}, pages 922--928. IEEE, 2015.

\bibitem{oord2016pixel}
A.~v.~d. Oord, N.~Kalchbrenner, and K.~Kavukcuoglu.
\newblock Pixel recurrent neural networks.
\newblock {\em arXiv preprint arXiv:1601.06759}, 2016.

\bibitem{oord2017parallel}
A.~v.~d. Oord, Y.~Li, I.~Babuschkin, K.~Simonyan, O.~Vinyals, K.~Kavukcuoglu,
  G.~v.~d. Driessche, E.~Lockhart, L.~C. Cobo, F.~Stimberg, et~al.
\newblock Parallel wavenet: Fast high-fidelity speech synthesis.
\newblock {\em arXiv preprint arXiv:1711.10433}, 2017.

\bibitem{pontes2017compact}
J.~K. Pontes, C.~Kong, A.~Eriksson, C.~Fookes, S.~Sridharan, and S.~Lucey.
\newblock Compact model representation for 3d reconstruction.
\newblock {\em arXiv preprint arXiv:1707.07360}, 2017.

\bibitem{pontes2017image2mesh}
J.~K. Pontes, C.~Kong, S.~Sridharan, S.~Lucey, A.~Eriksson, and C.~Fookes.
\newblock Image2mesh: A learning framework for single image 3d reconstruction.
\newblock {\em arXiv preprint arXiv:1711.10669}, 2017.

\bibitem{qi2017frustum}
C.~R. Qi, W.~Liu, C.~Wu, H.~Su, and L.~J. Guibas.
\newblock Frustum pointnets for 3d object detection from rgb-d data.
\newblock {\em arXiv preprint arXiv:1711.08488}, 2017.

\bibitem{qi2017pointnet}
C.~R. Qi, H.~Su, K.~Mo, and L.~J. Guibas.
\newblock Pointnet: Deep learning on point sets for 3d classification and
  segmentation.
\newblock {\em CVPR}, 2017.

\bibitem{qi2017pointnet++}
C.~R. Qi, L.~Yi, H.~Su, and L.~J. Guibas.
\newblock Pointnet++: Deep hierarchical feature learning on point sets in a
  metric space.
\newblock In {\em Advances in Neural Information Processing Systems}, pages
  5099--5108, 2017.

\bibitem{ramachandran2017fast}
P.~Ramachandran, T.~L. Paine, P.~Khorrami, M.~Babaeizadeh, S.~Chang, Y.~Zhang,
  M.~A. Hasegawa-Johnson, R.~H. Campbell, and T.~S. Huang.
\newblock Fast generation for convolutional autoregressive models.
\newblock {\em arXiv preprint arXiv:1704.06001}, 2017.

\bibitem{riegler2017octnet}
G.~Riegler, A.~O. Ulusoy, and A.~Geiger.
\newblock Octnet: Learning deep 3d representations at high resolutions.
\newblock In {\em Proceedings of the IEEE Conference on Computer Vision and
  Pattern Recognition}, volume~3, 2017.

\bibitem{salimans2017pixelcnn++}
T.~Salimans, A.~Karpathy, X.~Chen, and D.~P. Kingma.
\newblock Pixelcnn++: Improving the pixelcnn with discretized logistic mixture
  likelihood and other modifications.
\newblock {\em arXiv preprint arXiv:1701.05517}, 2017.

\bibitem{sengupta2017sfsnet}
S.~Sengupta, A.~Kanazawa, C.~D. Castillo, and D.~Jacobs.
\newblock Sfsnet: Learning shape, reflectance and illuminance of faces in the
  wild.
\newblock {\em arXiv preprint arXiv:1712.01261}, 2, 2017.

\bibitem{sharma2016vconv}
A.~Sharma, O.~Grau, and M.~Fritz.
\newblock Vconv-dae: Deep volumetric shape learning without object labels.
\newblock In {\em ECCV}, 2016.

\bibitem{shen2017neighbors}
Y.~Shen, C.~Feng, Y.~Yang, and D.~Tian.
\newblock Neighbors do help: Deeply exploiting local structures of point
  clouds.
\newblock {\em arXiv preprint arXiv:1712.06760}, 2017.

\bibitem{stets2017visualization}
J.~D. Stets, Y.~Sun, W.~Corning, and S.~W. Greenwald.
\newblock Visualization and labeling of point clouds in virtual reality.
\newblock In {\em SIGGRAPH Asia 2017 Posters}, 2017.

\bibitem{su2018splatnet}
H.~Su, V.~Jampani, D.~Sun, S.~Maji, E.~Kalogerakis, M.-H. Yang, and J.~Kautz.
\newblock Splatnet: Sparse lattice networks for point cloud processing.
\newblock In {\em Proceedings of the IEEE Conference on Computer Vision and
  Pattern Recognition}, pages 2530--2539, 2018.

\bibitem{sun2018x}
Y.~Sun, S.~N.~R. Kantareddy, R.~Bhattacharyya, and S.~E. Sarm.
\newblock X-vision: An augmented vision tool with real-time sensing ability in
  tagged environments.
\newblock {\em arXiv preprint arXiv:1806.00567}, 2018.

\bibitem{sun2018im2avatar}
Y.~Sun, Z.~Liu, Y.~Wang, and S.~E. Sarma.
\newblock Im2avatar: Colorful 3d reconstruction from a single image.
\newblock {\em arXiv preprint arXiv:1804.06375}, 2018.

\bibitem{tan2017variational}
Q.~Tan, L.~Gao, Y.-K. Lai, and S.~Xia.
\newblock Variational autoencoders for deforming 3d mesh models.
\newblock {\em arXiv preprint arXiv:1709.04307}, 2017.

\bibitem{tatarchenko2017octree}
M.~Tatarchenko, A.~Dosovitskiy, and T.~Brox.
\newblock Octree generating networks: Efficient convolutional architectures for
  high-resolution 3d outputs.
\newblock In {\em ICCV}, 2017.

\bibitem{te2018rgcnn}
G.~Te, W.~Hu, Z.~Guo, and A.~Zheng.
\newblock Rgcnn: Regularized graph cnn for point cloud segmentation.
\newblock {\em arXiv preprint arXiv:1806.02952}, 2018.

\bibitem{van2016wavenet}
A.~Van Den~Oord, S.~Dieleman, H.~Zen, K.~Simonyan, O.~Vinyals, A.~Graves,
  N.~Kalchbrenner, A.~W. Senior, and K.~Kavukcuoglu.
\newblock Wavenet: A generative model for raw audio.
\newblock In {\em SSW}, 2016.

\bibitem{van2016conditional}
A.~van~den Oord, N.~Kalchbrenner, L.~Espeholt, O.~Vinyals, A.~Graves, et~al.
\newblock Conditional image generation with pixelcnn decoders.
\newblock In {\em NIPS}, 2016.

\bibitem{vaswani2017attention}
A.~Vaswani, N.~Shazeer, N.~Parmar, J.~Uszkoreit, L.~Jones, A.~N. Gomez,
  {\L}.~Kaiser, and I.~Polosukhin.
\newblock Attention is all you need.
\newblock In {\em Advances in Neural Information Processing Systems}, pages
  5998--6008, 2017.

\bibitem{wang2018pixel2mesh}
N.~Wang, Y.~Zhang, Z.~Li, Y.~Fu, W.~Liu, and Y.-G. Jiang.
\newblock Pixel2mesh: Generating 3d mesh models from single rgb images.
\newblock {\em arXiv preprint arXiv:1804.01654}, 2018.

\bibitem{wang2017cnn}
P.-S. Wang, Y.~Liu, Y.-X. Guo, C.-Y. Sun, and X.~Tong.
\newblock O-cnn: Octree-based convolutional neural networks for 3d shape
  analysis.
\newblock {\em ACM Transactions on Graphics (TOG)}, 36(4):72, 2017.

\bibitem{wang2018dynamic}
Y.~Wang, Y.~Sun, Z.~Liu, S.~E. Sarma, M.~M. Bronstein, and J.~M. Solomon.
\newblock Dynamic graph cnn for learning on point clouds.
\newblock {\em arXiv preprint arXiv:1801.07829}, 2018.

\bibitem{wu2016learning}
J.~Wu, C.~Zhang, T.~Xue, B.~Freeman, and J.~Tenenbaum.
\newblock Learning a probabilistic latent space of object shapes via 3d
  generative-adversarial modeling.
\newblock In {\em Advances in Neural Information Processing Systems}, pages
  82--90, 2016.

\bibitem{wu20153d}
Z.~Wu, S.~Song, A.~Khosla, F.~Yu, L.~Zhang, X.~Tang, and J.~Xiao.
\newblock 3d shapenets: A deep representation for volumetric shapes.
\newblock In {\em Proceedings of the IEEE conference on computer vision and
  pattern recognition}, pages 1912--1920, 2015.

\bibitem{xiang2014beyond}
Y.~Xiang, R.~Mottaghi, and S.~Savarese.
\newblock Beyond pascal: A benchmark for 3d object detection in the wild.
\newblock In {\em WACV}, 2014.

\bibitem{yang2018foldingnet}
Y.~Yang, C.~Feng, Y.~Shen, and D.~Tian.
\newblock Foldingnet: Point cloud auto-encoder via deep grid deformation.
\newblock In {\em Proc. IEEE Conf. on Computer Vision and Pattern Recognition
  (CVPR)}, volume~3, 2018.

\bibitem{yi2017syncspeccnn}
L.~Yi, H.~Su, X.~Guo, and L.~J. Guibas.
\newblock Syncspeccnn: Synchronized spectral cnn for 3d shape segmentation.
\newblock In {\em CVPR}, pages 6584--6592, 2017.

\bibitem{yu2018pu}
L.~Yu, X.~Li, C.-W. Fu, D.~Cohen-Or, and P.-A. Heng.
\newblock Pu-net: Point cloud upsampling network.
\newblock In {\em Proceedings of the IEEE Conference on Computer Vision and
  Pattern Recognition}, pages 2790--2799, 2018.

\bibitem{yuan2018pcn}
W.~Yuan, T.~Khot, D.~Held, C.~Mertz, and M.~Hebert.
\newblock Pcn: Point completion network.
\newblock In {\em 2018 International Conference on 3D Vision (3DV)}, pages
  728--737. IEEE, 2018.

\bibitem{zhang2018self}
H.~Zhang, I.~Goodfellow, D.~Metaxas, and A.~Odena.
\newblock Self-attention generative adversarial networks.
\newblock {\em arXiv preprint arXiv:1805.08318}, 2018.

\end{thebibliography}
